\documentclass[letterpaper]{article} % DO NOT CHANGE THIS
\usepackage{aaai2027}
\usepackage{helvet}  % DO NOT CHANGE THIS
\usepackage{courier}  % DO NOT CHANGE THIS
\usepackage[hyphens]{url}  % DO NOT CHANGE THIS
\usepackage{graphicx} % DO NOT CHANGE THIS
\usepackage{natbib}  % DO NOT CHANGE THIS AND DO NOT ADD ANY OPTIONS TO IT
\usepackage{caption} % DO NOT CHANGE THIS AND DO NOT ADD ANY OPTIONS TO IT
\usepackage{algorithm}
\usepackage{algorithmic}

\usepackage{newfloat}
\usepackage{listings}
\DeclareCaptionStyle{ruled}{labelfont=normalfont,labelsep=colon,strut=off} % DO NOT CHANGE THIS
\floatstyle{ruled}
\newfloat{listing}{tb}{lst}{}
\floatname{listing}{Listing}
\usepackage{amsmath}
\usepackage{amssymb}
\usepackage{mathtools}
\usepackage{amsthm}
\usepackage{etoolbox} % for \AtEndEnvironment hooks
\usepackage{subcaption}

\theoremstyle{plain}
\newtheorem{theorem}{Theorem}
\newtheorem{proposition}[theorem]{Proposition}

\theoremstyle{definition}

\newtheorem{assumption}[theorem]{Assumption}
\newtheorem{example}{Example}
\newtheorem{condition}{Condition}

\theoremstyle{remark}
\newtheorem{remark}[theorem]{Remark}

\AtEndEnvironment{example}{\hfill$\blacksquare$}

\usepackage{enumitem}
\usepackage{mathrsfs}
\usepackage{xcolor}
\usepackage{multirow}
\usepackage{booktabs} % For \toprule, \midrule, \bottomrule
\usepackage{graphicx} % For resizing if needed
\usepackage{adjustbox} % Optional: scaling wide tables
\usepackage{caption}   % For better caption formatting
\usepackage{makecell} % For line breaks inside table

\newcommand{\D}{\Omega}
\def\*#1{\boldsymbol{#1}}
\def\1#1{\mathcal{#1}}
\def\2#1{\mathscr{#1}}
\def\3#1{\mathbb{#1}}

\definecolor{betterred}{RGB}{228,26,28}
\usepackage{tikz}
\usetikzlibrary{arrows.meta}
\tikzset{%
  >={Latex[width=1.5mm,length=2mm]},
  vertex/.style={draw,circle,inner sep=0mm,semithick,minimum width=4mm},
  bidir/.style={<->,dashed, line width=0.25mm},
  bidirdrop/.style={<->,dashed, line width=0.5mm, betterred!80},
  dir/.style={->, line width=0.25mm},
  node distance=1cm,
  font=\scriptsize\sffamily
}

\title{Differentiable Causal Discovery for Singular Linear Models under Confounding}

\author{
Mujin Zhou$^1$ \quad
Ignavier Ng$^2$ \quad
Junzhe Zhang$^3$
}

\affiliations{
$^1$University of Illinois Urbana-Champaign, \texttt{mujinz2@illinois.edu} \\
$^2$Carnegie Mellon University, \texttt{ignavierng@cmu.edu} \\
$^3$Syracuse University, \texttt{jzhan403@syr.edu}
}

\begin{document}

\maketitle

\begin{abstract}
Score-based causal discovery in the presence of unobserved confounders requires both a consistent scoring criterion and an efficient search over graph structures. Linear causal models with correlated errors are naturally represented by acyclic directed mixed graphs (ADMGs), whose induced model families include singular statistical models for which the standard Bayesian Information Criterion is inconsistent. We develop a score-based framework for causal discovery over linear Gaussian ADMGs grounded in singular learning theory. We first verify that the conditions for consistency of the Widely Applicable Bayesian Information Criterion (WBIC) are satisfied by the linear causal model class. We then propose a scalable approximation of the WBIC via Automatic Differentiation Variational Inference (ADVI). Finally, we introduce a differentiable search over ADMGs through Gumbel-Softmax relaxations of both directed and bidirected adjacency. Experiments on synthetic and real-world benchmarks demonstrate improvements over the BIC-scored baseline on both singular and generic ADMGs under unobserved confounding.
\end{abstract}

\section{Introduction}
Learning the causal structure underlying a particular phenomenon from data is a fundamental problem across the data sciences. One of the common approaches in causal discovery models the underlying system as a causal model represented by a directed acyclic causal graph, where nodes denote random variables (measured or latent) and directed edges denote causal effects from tails to arrowheads \cite{pearl:2k,spirtes2000causation,peters2017elements,bareinboim2020pearl}. The task is then to piece together the constraints found in the data (and implied by the underlying, unobserved causal system) to infer the corresponding causal graph.

There are various types of statistical constraints imposed by the underlying causal system on the observed data drawn from the observational distribution $P(\*V)$. In the linear causal models, an $d$-separation between nodes in a causal graph induces a corresponding conditional independence between variables in $\*V$, which in turn corresponds to a zero partial correlation. The reverse implication, i.e.\ that each conditional independence in data implies a corresponding $d$-separation in the underlying causal graph (known as faithfulness), serves as a statistically testable constraint to narrow the class of compatible graphs \cite{pearl:88a,meek:95,uhler2013geometry,zhang2006causal}. This is the cornerstone assumption for a plethora of structure learning algorithms \cite{verma1990causal,glymour2014discovering,spirtes1991pc,spirtes2000causation}. When structural functions are linear and exogenous noises are independent, $d$-separation statements capture \textit{all} testable constraints implied by the underlying causal model \cite{verma1990causal}.

This is not the case in systems involving unobserved confounding. In linear causal models, unobserved confounders induce correlations among the exogenous noise terms of the structural equations, and the resulting model is naturally represented by an \emph{Acyclic Directed Mixed Graph} (ADMG) \cite{richardson2002ancestral}, whose bidirected edges encode latent common causes alongside the directed causal links. Such models impose statistical constraints beyond conditional independence: most notably, they satisfy \textit{tetrad constraints} \cite{sullivant2010trek}, polynomial equalities on the observable covariance matrix that can distinguish causal structures which conditional-independence tests alone cannot. Because these constraints are invisible to independence tests yet reflected in the likelihood, they motivate \emph{score-based} causal discovery \citep{chickering2002optimal}, which ranks structures by a global fit criterion rather than a sequence of independence decisions.

In the linear Gaussian setting with latent variables, however, the induced family of observational distributions is generally \textit{not} a curved exponential family \cite{geiger2001stratified}, breaking a key assumption behind standard scores. The Bayesian Information Criterion (BIC), derived as an asymptotic approximation to the model posterior for curved exponential families \cite{haughton1988choice,schwarz1978estimating}, is therefore no longer directly applicable: \citet{drton2009likelihood} shows that the Fisher information of linear Gaussian models with correlated errors can be singular at certain parameter configurations, so the standard asymptotics underlying the BIC no longer hold \cite{drton2017structure}.

Building on Watanabe's singular learning theory \cite{watanabe2009algebraic}, the Widely Applicable Bayesian Information Criterion (WBIC) \cite{watanabe2013widely} provides a principled generalization of the BIC to singular models. The WBIC replaces the standard model dimension penalty with a data-dependent approximation of the \textit{real log-canonical threshold} (RLCT), a birational invariant that characterizes the singularity structure and provides a refined measure of effective model complexity that remains valid even when the Fisher information is singular. Unlike the BIC, the WBIC remains a consistent model selection criterion for singular models, making it a natural candidate for scoring linear Gaussian causal models with unobserved confounders represented as ADMGs. However, the exact computation of the WBIC requires posterior expectations that are generally intractable for models of practical interest, necessitating scalable approximate inference.

A consistent score is only half the problem; how the space of structures is searched has its own history and limitations. Early structure learning approaches for the linear Gaussian setting with latent variables, including the FCI algorithm \cite{spirtes2000causation} and its variants \cite{colombo2012learning,zhang2008completeness}, operate by testing conditional independencies. Score-based approaches \cite{chickering2002optimal,hauser2012characterization} have largely focused on the fully observed case, where the BIC provides a consistent and decomposable score. More recently, continuous relaxation methods such as NOTEARS \cite{zheng2018dags} have recast the combinatorial problem of DAG search as a continuous constrained optimization, enabling gradient-based search over graph structures. \citet{bhattacharya2021differentiable} extend this continuous-optimization approach to ADMGs, searching over bow-free graphs, those in which no pair of variables is joined by both a directed and a bidirected edge, a restriction that guarantees the model is globally identifiable. This search still scores structures with the BIC, however, and so fails on the singular model classes that latent confounding induces, as we
demonstrate in Example~\ref{ex:drton}. Extending score-based causal discovery to linear ADMGs thus requires addressing two coupled challenges: (i) computing a consistent score for singular model classes, and (ii) searching efficiently over the space of ADMG equivalence classes.

Our goal in this paper is to address both challenges. We propose a score-based approach for causal discovery in linear Gaussian ADMG models. More specifically, our contributions are as follows. (1) We establish theoretical guarantees for WBIC consistency in the linear Gaussian ADMG setting by verifying the four fundamental conditions of Watanabe's singular learning theory \cite{watanabe2013widely,watanabe2009algebraic}: a compact parameter space, a factorized prior, analytic $L^s$-integrability of the log-density, and relatively finite variance. Together with the assumption that the real log-canonical threshold (RLCT) orders linear ADMG models consistently, these conditions yield global large-sample consistency of the WBIC for linear causal models with unobserved confounders. (2) We develop a scalable ADVI-based approximation of the WBIC \cite{kucukelbir2017automatic} that replaces the intractable tempered posterior expectation with a mean-field Gaussian variational surrogate. (3) We propose a differentiable search over ADMG equivalence classes that relaxes both directed and bidirected adjacency with Gumbel-Softmax and reports the recovered Markov equivalence class as a PAG. Due to space constraints, all the proofs and details of the experimental setups are provided in the appendix.

\section{Preliminaries}
We use capital letters for random variables ($X$), lowercase for their values ($x$),
and bold for sets ($\*V$) and their values ($\*v$). The domain of $X$ is $\D_X$, and
$P(\*V)$ is the distribution over $\*V$, with $P(\*v)$ shorthand for $P(\*V=\*v)$.

\paragraph{Structural Causal Models.}
We work in the language of \textit{structural causal models} (SCMs) \cite{pearl:2k}.
A linear semi-Markovian SCM $M$ over endogenous variables $\*V$ and exogenous
variables $\*U$ is a system of linear equations $V=\Lambda V+U$ for $V\in\*V$, where
$\Lambda$ collects the structural coefficients, with $\lambda_{ij}=0$ indicating no
direct effect of $V_j$ on $V_i$. The disturbances $\*U$ are centered multivariate
Gaussian with \textit{error covariance} $\1E$: the diagonal $\omega_{ii}$ are error
variances and the off-diagonal $\omega_{ij}$ encode disturbance covariances, so
$\omega_{ij}=0$ means $V_i,V_j$ share no unobserved common cause. We restrict to
acyclic models, so $\Lambda$ is lower triangular, and the induced endogenous
covariance is $\Sigma=(I-\Lambda)^{-1}\1E(I-\Lambda)^{-\top}$. Without loss of
generality all variables are standardized to unit variance. For variables $X,Y$ and a
conditioning set $Z$, $\sigma_{YX}$ is their covariance, $\sigma_{YX\cdot Z}$ the
partial covariance given $Z$, and $R_{YX\cdot Z}$ the regression coefficient of $Y$ on
$X$ adjusting for $Z$. Causal quantities of interest are entries of $\Lambda$ or
functions thereof, and identifiability reduces to whether they are uniquely
recoverable from $\Sigma$.

\paragraph{Causal Diagrams and ADMGs.}
The structural assumptions of $M$ are encoded in its \textit{causal graph}
$\1G=(\*V,\*D,\*B)$: vertices $\*V$ are the endogenous variables, directed edges
$\*D$ the nonzero entries of $\Lambda$, and bidirected edges $\*B$ the nonzero
off-diagonal entries of $\1E$. We consider \textit{Acyclic Directed Mixed Graphs}
(ADMGs) \cite{richardson2002ancestral}, which carry directed ($\rightarrow$) and
bidirected ($\leftrightarrow$) edges but no directed cycles. A missing directed edge
encodes an \textit{exclusion restriction} (no direct effect); a missing bidirected
edge encodes an \textit{independence restriction} (no unobserved common cause). When
context is clear we treat coefficients and their graph edges interchangeably.

\paragraph{Bayesian Free Energy and the BIC.}
Given $n$ i.i.d.\ observations $\*D_n=\{\*v_1,\dots,\*v_n\}$, Bayesian model selection
ranks candidate classes $\1M$ by the \textit{Bayesian free energy} $F_n(\1M)$, the
negative log evidence
$F_n(\1M)=-\log\int\prod_{i=1}^{n}p(\*v_i\mid\theta,\1M)\,\varphi(\theta\mid\1M)\,d\theta$,
where $\theta$ collects the free parameters (the nonzero entries of $\Lambda$ and
$\1E$) and $\varphi$ is a prior; the model minimizing $F_n$ is preferred. For
\textit{regular} models—curved exponential families with nonsingular Fisher
information—a Laplace approximation gives the \textit{Bayesian Information Criterion}
(BIC) as leading order,
\begin{align}
    F_n(\1M) = -\log p(\*D_n \mid \hat\theta, \1M) + \frac{d}{2}\log n + O_p(1),
    \label{eq:bic}
\end{align}
with MLE $\hat\theta$ and $d=\dim(\theta)$, recovering
$\mathrm{BIC}(\1M)=-\log p(\*D_n\mid\hat\theta,\1M)+(d/2)\log n$
\cite{schwarz1978estimating,haughton1988choice}. This relies on regularity: for
singular classes such as linear Gaussian ADMGs with correlated errors the Fisher
information degenerates, the $(d/2)\log n$ penalty no longer reflects effective
complexity, and the free energy must be approached through singular learning theory
\cite{watanabe2009algebraic} and the WBIC \cite{watanabe2013widely}.

\paragraph{Score-Based Causal Discovery.}
Score-based discovery scores each candidate structure by the free energy or BIC above and searches for the highest-scoring one. Classically this search is combinatorial, ranging over DAGs by greedy moves through Markov equivalence classes \cite{chickering2002optimal}, with acyclicity enforced structurally by the search operators. Continuous-relaxation methods instead recast the search as smooth optimization over a weighted adjacency matrix, turning acyclicity from a combinatorial property into a differentiable constraint. NOTEARS \cite{zheng2018dags} does so through the trace function $h(W)=\operatorname{tr}[\exp(W\circ W)]-p$, with $\circ$ the Hadamard product and $p$ the number of variables, which vanishes exactly when $W$ is acyclic and can be imposed as a penalty so the score is optimized by gradient methods. These methods score with the BIC, however, which is inconsistent for the singular model classes that latent confounding induces.

\section{Challenges of Singular Models}
Linear Gaussian SCMs over ADMGs are not, in general, regular statistical models. Write $\phi_\1G$ for the parametrization that maps the structural parameters $(\Lambda, \1E)$ of a graph $\1G$ to the observed covariance matrix $\Sigma$; identifiability of $\1G$ is then injectivity of $\phi_\1G$. When unobserved confounders induce correlated errors, $\phi_\1G$ can fail to be injective on a lower-dimensional subset of the parameter space. The model is then \textit{generically} but not \textit{globally} identifiable, and $\phi_\1G$ is singular at such points. There, a positive-dimensional set of parameters maps to the same distribution, so the parameters are unconstrained along these directions and the model has fewer than $d$ effective degrees of freedom. Its \emph{effective} dimension is thus strictly smaller than the nominal count $d$. The BIC expansion of Eq.~\eqref{eq:bic} no longer applies: the $\tfrac{d}{2}\log n$ term overcounts complexity, and the right replacement is the real log-canonical threshold (RLCT) that governs the WBIC \cite{drton2009likelihood,drton2017structure}.

\begin{example}\label{ex:drton}
Consider the five-node ADMG $\1G$ in Fig.~\ref{fig:drton}, with directed edges $1\to2\to3\to4\to5$ and bidirected edges $1\leftrightarrow4$, $1\leftrightarrow5$, $2\leftrightarrow4$, and $3\leftrightarrow5$ \citep{drton2011global}. Take the structural coefficients $\lambda_{12}=3$, $\lambda_{23}=-\tfrac12$, $\lambda_{34}=\lambda_{45}=1$, and let $\1E$ have unit diagonal scaled to $2$ with $\omega_{14}=\omega_{15}=\omega_{24}=\omega_{35}=1$. At this point the map $\phi_\1G$ is not injective and its image has a singularity: $\1G$ is generically but not globally identifiable. This singularity is precisely the regime in which the BIC adopts the wrong effective dimension \citep{drton2011global}. Replacing $\lambda_{23}=-\tfrac12$ by $\lambda_{23}=+\tfrac12$ removes the singularity and yields a regular model at which standard asymptotics hold.

To see in practice, we simulate $n=1000$ samples from the model of Example~\ref{ex:drton} and score, with the standard BIC, the true graph $\1G$ against the sparser submodel $\1G'$ obtained by deleting the bidirected edge $1\leftrightarrow5$, the edge whose coefficient is unidentified along the singularity's flat direction. The BIC row of Table~\ref{tab:selection} (averaged over $R=30$ replicates at $n=1000$, in the free-energy convention, so lower is better) tells the story. At the regular point ($\lambda_{23}=+\tfrac12$) the edge is identified, deleting it genuinely degrades the fit, and the BIC ranks the true $\1G$ decisively best ($16224.5$ versus $16558.0$). At the singular point ($\lambda_{23}=-\tfrac12$) the edge becomes redundant ($\1G$ and $\1G'$ induce the same distribution, as we verify explicitly in appendix, Prop.~\ref{prop:equiv-ex1}, on distributional equivalence), yet the BIC, charging the full nominal dimension $d$ through $\tfrac{d}{2}\log n$ penalty, prefers the sparser $\1G'$ ($16023.6$ versus $16029.3$) and thereby discards the ground-truth graph. The failure is structural rather than statistical: at the singularity, the effective dimension is strictly below $d$, so the nominal-dimension penalty overcharges $\1G$. Recovering the true graph demands a score whose complexity penalty tracks \emph{effective} rather than nominal dimension.
\end{example}

\begin{figure}[t]
\centering
\begin{tikzpicture}[font=\scriptsize\sffamily]
  \node[vertex] (n1) at (0,0)      {1};
  \node[vertex] (n2) at (1.1,0.85) {2};
  \node[vertex] (n3) at (2.2,1.7)  {3};
  \node[vertex] (n4) at (3.3,0.85) {4};
  \node[vertex] (n5) at (4.7,0.85) {5};
  \draw[dir] (n1) -- (n2);
  \draw[dir] (n2) -- (n3);
  \draw[dir] (n3) -- (n4);
  \draw[dir] (n4) -- (n5);
  \draw[bidir] (n1) to[bend right=22] (n4);
  \draw[bidirdrop] (n1) to[bend right=38] (n5);
  \draw[bidir] (n2) to[bend right=18] (n4);
  \draw[bidir] (n3) to[bend left=26]  (n5);
\end{tikzpicture}
\caption{The acyclic mixed graph of \citeauthor{drton2011global}'s Example~1 \citep{drton2011global}; solid edges are directed and dashed edges bidirected. At the parameter point of Example~\ref{ex:drton} the parametrization $\phi_\1G$ is non-injective, so the induced Gaussian model is singular. The highlighted bidirected edge $1\leftrightarrow5$ (red) is redundant at this singular point: it can be removed without changing the induced distribution, yielding the submodel $\1G'$.}
\label{fig:drton}
\end{figure}

\begin{table*}[t]
\centering
\setlength{\tabcolsep}{12pt}
\begin{tabular}{l cc cc}
\toprule
 & \multicolumn{2}{c}{Regular ($\lambda_{23}=+1/2$)}
 & \multicolumn{2}{c}{Singular ($\lambda_{23}=-1/2$)} \\
\cmidrule(lr){2-3}\cmidrule(lr){4-5}
Criterion
 & \shortstack{$\1G$ (Fig.~\ref{fig:drton})}
 & \shortstack{$\1G'$ (Fig.~\ref{fig:drton} w/o $1\!\leftrightarrow\!5 $)}
 & \shortstack{$\1G$ (Fig.~\ref{fig:drton})}
 & \shortstack{$\1G'$ (Fig.~\ref{fig:drton} w/o $1\!\leftrightarrow\!5 $)} \\
\midrule
BIC  & $\mathbf{16224.54 \pm 116.62}$ & $16558.03 \pm 120.82$ & $16029.31 \pm 111.43$ & $16023.58 \pm 111.15$ \\
WBIC & $\mathbf{8017.56 \pm 55.28}$   & $8282.18 \pm 61.42$   & $\mathbf{8012.94 \pm 55.94}$ & $\mathbf{8013.93 \pm 57.28}$ \\
ADVI & $\mathbf{8079.24 \pm 58.21}$   & $8248.41 \pm 60.54$   & $\mathbf{7981.55 \pm 55.77}$ & $\mathbf{7981.19 \pm 55.68}$ \\
\bottomrule
\end{tabular}
\caption{Selection scores (mean $\pm$ s.d.\ over $R=30$ replicates, $n=1000$) for the true graph $\1G$ and its sparser submodel $\1G'$; lower is better (free-energy convention) and \textbf{bold} marks the regular-regime winner. At the regular point, every criterion ranks $\1G$ decisively best (margins $\sim$170--330). At the singular point $\1G$ and $\1G'$ are near-equivalent and the gap all but vanishes: only the BIC tilts toward the sparser $\1G'$, while the WBIC and its  ADVI approximation leave the two essentially tied.}
\label{tab:selection}
\end{table*}

\subsection{The Widely Applicable BIC}
Watanabe's singular learning theory replaces the nominal-dimension penalty by the \emph{real log-canonical threshold} (RLCT) $\lambda$, a birational invariant that measures how fast the likelihood concentrates near the optimal parameter set. The RLCT plays the role of an \emph{effective} half-dimension of the model, coinciding with $d/2$ only in the regular case \cite{watanabe2009algebraic}. Writing $L_n(\theta) = -\tfrac{1}{n}\sum_{i=1}^{n}\log p(\*v_i \mid \theta, \1M)$ for the empirical negative log-likelihood, the WBIC \cite{watanabe2013widely} estimates the Bayesian free energy $F_n(\1M)$ by a single tempered posterior expectation,
\begin{align}
\mathrm{WBIC}(\1M) = \mathbb{E}^{\beta}_{\theta}\!\left[\, n L_n(\theta) \,\right], \qquad \beta = \frac{1}{\log n},
\label{eq:wbic}
\end{align}
where $\mathbb{E}^{\beta}_{\theta}[g] = \int g(\theta)\,\pi_n^{\beta}(\theta)\,d\theta$ is the expectation under the posterior distribution $\pi_n^{\beta}(\theta) \propto e^{-n\beta L_n(\theta)}\varphi(\theta)$ tempered at inverse temperature $\beta$. The value $\beta = 1/\log n$ is the crucial choice: Watanabe shows that the free energy equals this posterior-averaged log-likelihood at exactly this single temperature, so one tempered expectation, rather than a thermodynamic integration over many, already recovers the RLCT penalty. Under the fundamental conditions reviewed below, the WBIC admits the asymptotic expansion
\begin{align}
\mathrm{WBIC}(\1M) = n L_n(\hat\theta) + \lambda\,\log n + O_p\!\big(\sqrt{\log n}\,\big),
\label{eq:wbic-expansion}
\end{align}
where $\lambda > 0$ is the RLCT of $\1M$ at the population-optimal parameter \cite{watanabe2013widely}. For regular models $\lambda = d/2$ and Eq.~\eqref{eq:wbic-expansion} reduces to the BIC of Eq.~\eqref{eq:bic}; at a singularity $\lambda < d/2$, the WBIC charges strictly less than the nominal penalty, exactly the singular regime of Example~\ref{ex:drton} demands.

For a linear Gaussian ADMG the likelihood is determined by the model-implied covariance $\Sigma(\theta) = (I-\Lambda)^{-1}\,\1E\,(I-\Lambda)^{-\top}$, where $\theta = (\Lambda, \1E)$ ranges over the graph-restricted support fixed by $\1G$ (zeros off $\*D$ in $\Lambda$ and off $\*B$ in $\1E$). The next proposition instantiates Eq.~\eqref{eq:wbic-expansion} for this class.

\begin{proposition}[WBIC for linear Gaussian ADMGs]\label{prop:wbic-admg}
Let $\1G$ be an ADMG on $p$ vertices and let $S_n$ denote the sample covariance of $n$ i.i.d.\ observations. The empirical negative log-likelihood of $\1M(\1G)$ is
\begin{align}
&n L_n(\theta) \notag \\
&= \frac{n}{2}\Big[\, p\log 2\pi + \log\det\Sigma(\theta) + \operatorname{tr}\!\big(\Sigma(\theta)^{-1} S_n\big) \Big],\label{eq:gauss-loss}
\end{align}
and, writing $\hat\theta_\1G$ for its minimizer over the graph-restricted support, the WBIC of $\1G$ satisfies
\begin{align}
\mathrm{WBIC}(\1G) = n L_n(\hat\theta_\1G) + \lambda_\1G\,\log n + O_p\!\big(\sqrt{\log n}\,\big),
\label{eq:wbic-admg}
\end{align}
where $\lambda_\1G$ is the RLCT of $\1M(\1G)$ at its best-fitting parameter.
\end{proposition}

Eq.~\eqref{eq:gauss-loss} is the Gaussian log-likelihood under the ADMG parametrization, and Eq.~\eqref{eq:wbic-admg} is Eq.~\eqref{eq:wbic-expansion} applied to $\1M(\1G)$; the only substantive claim is that $\1M(\1G)$ meets the hypotheses of Eq.~\eqref{eq:wbic-expansion}, which we verify next.

Expansion Eq.~\eqref{eq:wbic-expansion} holds whenever the triple $(q, p(\cdot\mid\theta), \varphi)$ of true density, model, and prior satisfies Watanabe's fundamental conditions \cite{watanabe2013widely}. These are mild regularity requirements that make the free-energy geometry well defined: analyticity so the singularities can be resolved, compactness and a smooth prior so the defining integral is controlled, and a finite-variance bound tying the Kullback--Leibler divergence to the fluctuations of the log-likelihood. Writing $f(x,\theta) = \log\{q(x)/p(x\mid\theta)\}$ for the log-density ratio, $K(\theta) = \mathbb{E}_X[f(X,\theta)]$ for the Kullback--Leibler divergence (the expectation $\mathbb{E}_X$ taken under the true density $q$, i.e.\ $X \sim q$), $L^s(q)$ for the space of functions $s$-integrable under $q$, and $W$ for the parameter set, they read:
\begin{condition}[Compactness]\label{cond:compact}
$W \subset \mathbb{R}^d$ is compact with non-empty interior, and its boundary is the zero set of finitely many analytic functions.
\end{condition}
\begin{condition}[Prior factorization]\label{cond:prior}
$\varphi(\theta) = \varphi_1(\theta)\,\varphi_2(\theta)$ with $\varphi_1 \ge 0$ analytic and $\varphi_2 > 0$ of class $C^{\infty}$ (i.e.\ smooth, with derivatives of all orders).
\end{condition}
\begin{condition}[Analytic integrability]\label{cond:analytic}
There is an open $W' \supset W$ on which $\theta \mapsto f(x,\theta)$ is an $L^s(q)$-valued analytic function for some $s \ge 6$.
\end{condition}
\begin{condition}[Relatively finite variance]\label{cond:rfv}
There exist $\epsilon, c > 0$ such that $K(\theta) \ge c\,\mathbb{E}_X[f(X,\theta)^2]$ for all $\theta \in W_\epsilon = \{\theta \in W : K(\theta) \le \epsilon\}$.
\end{condition}

\begin{proposition}[Linear ADMGs are admissible]\label{prop:conditions}
For every ADMG $\1G$, the linear Gaussian model $\1M(\1G)$, restricted to a compact parameter set on which $\1E$ has eigenvalues in a fixed interval $[c_0, C_0] \subset (0,\infty)$ and equipped with any prior of the form in Condition~\ref{cond:prior}, satisfies Conditions \ref{cond:compact}--\ref{cond:rfv}. Consequently Eqs.~\eqref{eq:wbic-expansion} and \eqref{eq:wbic-admg} hold for $\1M(\1G)$.
\end{proposition}

The expansion Eq.~\eqref{eq:wbic-admg} reduces model selection to comparing $n L_n(\hat\theta_\1G) + \lambda_\1G \log n$ across graphs. Models that cannot represent the truth incur a linearly growing data-fit penalty; among those that can, the data-fit terms agree to $O_p(1)$ and the ranking is decided by the RLCTs $\lambda_\1G$. Consistency hinges on the RLCTs ordering the correctly-specified models in favour of the truth, which we adopt as an assumption.
\begin{assumption}[Global RLCT consistency]\label{asm:rlct}
Let $\1G^\star$ be the data-generating ADMG and $q$ the induced distribution. Among all ADMGs $\1G$ with $q \in \1M(\1G)$, the RLCT is minimized at $\1G^\star$: $\lambda_{\1G^\star} \le \lambda_\1G$, with equality only for $\1G$ distributionally equivalent to $\1G^\star$.
\end{assumption}

\begin{theorem}[Global consistency of the WBIC]\label{thm:consistency}
Suppose every candidate $\1M(\1G)$ satisfies Conditions \ref{cond:compact}--\ref{cond:rfv} (Prop.~\ref{prop:conditions}) and that Assumption~\ref{asm:rlct} holds. Then $\arg\min_\1G \mathrm{WBIC}(\1G)$ equals $\1G^\star$, up to distributional equivalence, with probability tending to $1$ as $n \to \infty$.
\end{theorem}
%Proofs of Prop.~\ref{prop:conditions} and Thm.~\ref{thm:consistency} are given in the appendix. 
Thm.~\ref{thm:consistency} is an asymptotic guarantee; to see it act on the concrete singular instance, we return to Example~\ref{ex:drton}.

\begin{example}[Example~\ref{ex:drton}, continued]\label{ex:drton-wbic}
The WBIC bears out this theory on the singular model of Example~\ref{ex:drton}. The WBIC row of Table~\ref{tab:selection} leaves $\1G$ and $\1G'$ essentially tied at the singular point, with the true $\1G$ still marginally ahead ($8012.9$ versus $8013.9$), precisely where the BIC fails, while agreeing with the BIC in the regular regime. The WBIC thus repairs the BIC's mis-scoring of the over-parameterized singular model and supplies the consistent score on which the differentiable search developed next is built. Its exact evaluation, however, requires the intractable tempered posterior of Eq.~\eqref{eq:wbic}, which we make scalable next.
\end{example}

\section{Scalable and Differentiable Causal Discovery}

WBIC gives a consistent score for comparing fixed ADMGs, but on its own it does not amount to a practical causal-discovery procedure. Evaluating it requires an expectation under a tempered posterior that is costly to recompute for every candidate graph, and the space of graphs is itself discrete. We tackle these two obstacles in turn: first we make WBIC scalable for a fixed ADMG through variational inference, and then we reuse the same approximation inside a continuous relaxation of the ADMG structure.

Let $\mathcal{G}=(\mathbf{V},D,B)$ be a fixed ADMG, where $D$ and $B$ denote its directed and bidirected edges. Its parameters are the associated edge weights $\theta=(\lambda,\omega)$, in which $\lambda_{ij}$ is the coefficient of the directed edge $V_j\to V_i\in D$ and $\omega$ collects the diagonal error variances together with the covariance weights of the bidirected edges $V_i\leftrightarrow V_j\in B$. The structure $\mathcal{G}$ therefore fixes $\theta$ completely: its directed and bidirected edges determine which weights are free, and every remaining coefficient and error covariance is held at zero. For a fixed graph $\1G$, recall that WBIC is the tempered-posterior expectation in Eq.~\ref{eq:wbic} with an inverse temperature $\beta=1/\log n$. Integrating this against the posterior gives,
\begin{align}
\mathrm{WBIC}(\mathcal{G}) = \int nL_n(\theta)\, \pi_n^\beta(\theta\mid\mathbf{D}_n,\mathcal{G}) \,d\theta,
\label{eq:wbic-posterior-expectation}
\end{align}
where the generalized tempered-posterior is
\begin{align}
\pi_n^\beta(\theta\mid\mathbf{D}_n,\mathcal{G}) \propto p(\mathbf{D}_n\mid\theta,\mathcal{G})^\beta\, \pi_\theta(\theta\mid\mathcal{G}).
\label{eq:tempered-posterior-fixed}
\end{align}
%The inverse temperature $\beta=1/\log n$ produces the singular-theoretic correction underlying WBIC: the posterior still concentrates around the optimal parameter set, while retaining the singular-volume behavior encoded by the RLCT.
One could estimate Eq.~\eqref{eq:wbic-posterior-expectation} by drawing samples from the generalized posterior with Metropolis--Hastings, but this is impractical inside structure learning, since it demands a separate Markov chain for every candidate graph. We instead approximate
$\pi_n^\beta(\theta\mid\mathbf{D}_n,\mathcal{G})$
with Automatic Differentiation Variational Inference (ADVI) \cite{kucukelbir2017automatic}.

The weights $\theta=(\lambda,\omega)$ are not free to vary arbitrarily: the directed coefficients are unrestricted, but the error covariance matrix must be symmetric positive definite. To let ADVI work in an unconstrained space, we reparameterize $\theta$ through $\zeta=T_{\mathcal{G}}(\theta)$, with $\zeta=(\zeta_\lambda,\zeta_L,\zeta_{\mathrm{diag}})\in\mathbb{R}^{d_{\mathcal{G}}}$. On the directed part this map is the identity, $\lambda_{ij}=\zeta_{\lambda,ij}$ for $V_j\to V_i\in D$, while the error covariance is built from a gated Cholesky factor $\Omega_\omega=LL^\top$ whose entries are
\begin{align}
L_{ii} &= \exp(\zeta_{\mathrm{diag},i}), \\
L_{ij} &= \mathbf{1}\{V_i\leftrightarrow V_j\in B\}\,\zeta_{L,ij},\quad i>j.
\label{eq:fixed-gated-cholesky}
\end{align}
This construction guarantees $\Omega_\omega\succ0$ automatically.%
\footnote{For a fixed ADMG, the Cholesky factor $L$
does not preserve the same zero pattern in the lower $\Omega$. Thus, the construction for fixed structure in
Eq.~\eqref{eq:fixed-gated-cholesky} introduces an
inherent bias. But in the continuous structure search,
no covariance zero pattern is fixed in advance: the structural parameters are continuous and are optimized jointly. The relaxation may still induce bias, but it adapts $LL^\top$ toward the covariance pattern supported by the data.}

We approximate the generalized posterior in these coordinates by a mean-field Gaussian,
\begin{align*}
q_\psi(\zeta) = \prod_{k\in\mathcal{I}} \mathcal{N}\!\left(\zeta_k;\mu_k,\sigma_k^2\right),
\end{align*}
where the index set $\mathcal{I}$ collects all free coordinates: the directed coordinates $\zeta_{\lambda,ij}$ for $(i,j)\in D$, the off-diagonal Cholesky coordinates $\zeta_{L,ij}$ for $i>j$ with $(i,j)\in B$, and the log-diagonal coordinates $\zeta_{\mathrm{diag},i}$ for $1\le i\le p$; the variational parameters are $\psi=(\mu,\log\sigma)$. Transforming back through $\theta=T_{\mathcal{G}}^{-1}(\zeta)$ carries this family to an approximation of the generalized posterior over $\theta$, whose density in $\zeta$ coordinates is $\pi_n^\beta(\zeta\mid\mathbf{D}_n,\mathcal{G}) \propto p(\mathbf{D}_n\mid T_{\mathcal{G}}^{-1}(\zeta),\mathcal{G})^\beta\, \pi_\zeta(\zeta\mid\mathcal{G})$.

We fit the variational parameters by maximizing the tempered evidence lower bound (ELBO),
\begin{equation}
\mathcal{L}_\beta(\psi;\mathcal{G}) = \mathbb{E}_{q_\psi(\zeta)}\!\left[ \beta\,\ell_n(\zeta;\mathbf{D}_n,\mathcal{G}) \right] - \mathrm{KL}\!\left( q_\psi(\zeta) \,\|\, \pi_\zeta(\zeta) \right).
\label{eq:elbo}
\end{equation}
The reparameterization $\zeta=\mu+\sigma\odot\varepsilon$ with $\varepsilon\sim\mathcal{N}(0,I)$ lets us form a Monte Carlo estimate of the gradient of Eq.~\eqref{eq:elbo} and maximize it by stochastic gradient ascent. We write $\psi^\star=\arg\max_\psi \mathcal{L}_{\beta}(\psi;\mathcal{G})$ for the resulting fit.

The fitted distribution $q_{\psi^\star}(\zeta)$ now stands in for the generalized posterior of Eq.~\eqref{eq:tempered-posterior-fixed}, and we estimate WBIC by averaging the loss over samples drawn from it,
\begin{align}
\widehat{\mathrm{WBIC}}(\mathcal{G}) &= \frac{1}{S}\sum_{s=1}^{S} nL_n(\theta^{(s)}), \label{eq:wbic-advi}\\
\theta^{(s)} &= T_{\mathcal{G}}^{-1}\!\left( \mu^\star+\sigma^\star\odot\varepsilon_s \right), \quad \varepsilon_s\sim\mathcal{N}(0,I). \nonumber
\end{align}
It is the single temperature $\beta=1/\log n$ that keeps this approximation compatible with singular learning theory. At this temperature, the effective sample size is $n\beta=n/\log n$, so the generalized posterior still concentrates on the optimal set, yet slowly enough that its leading behavior remains governed by the same singular geometry as WBIC itself. \footnote{In practice we evaluate the surrogate at a slightly colder temperature, $\beta = c/\log n$ with $c = 4$ (any $c \in [2,4]$ works well), as a heuristic: the mean-field family underestimates the posterior correlations along the singular set, and a colder $\beta$ suppresses this bias while keeping the over-parameterized graph distinguishable from its submodels. Theoretical results use the standard $\beta = 1/\log n$.}

\begin{proposition}[Consistency of the ADVI surrogate]\label{prop:advi}
Fix an ADMG $\1G$ satisfying Conditions \ref{cond:compact}--\ref{cond:rfv}. At temperature $\beta = 1/\log n$, the mean-field optimum $\psi^\star$ of Eq.~\eqref{eq:elbo} yields
\begin{align*}
\widehat{\mathrm{WBIC}}(\1G) = n L_n(\hat\theta_\1G) + \lambda_\1G^{\mathrm{vi}}\,\log n + o_p(\log n),
\end{align*}
where $\lambda_\1G^{\mathrm{vi}} > 0$ is the mean-field counterpart of the RLCT $\lambda_\1G$. If $\lambda^{\mathrm{vi}}$ preserves the ordering of Assumption~\ref{asm:rlct}, then selecting graphs by $\widehat{\mathrm{WBIC}}$ is globally consistent.
\end{proposition}
The surrogate thus inherits the structure of Eq.~\eqref{eq:wbic}: it reproduces the exact data-fit term and replaces the RLCT by a variational analogue whose leading $\log n$ coefficient controls model selection. Intuitively, the mean-field family matches the regular directions of the optimal set exactly, so $\lambda^{\mathrm{vi}}$ coincides with $\lambda_\1G$ whenever that set is a smooth manifold and can deviate only at genuine singular crossings. A proof is given in the appendix.

%Thm.~\ref{thm:consistency} and Prop.~\ref{prop:advi} are asymptotic; we close by confirming that the ADVI surrogate tracks the exact WBIC's singular-aware verdict on the instance of Example~\ref{ex:drton}.

\begin{example}[Example~\ref{ex:drton}, continued]\label{ex:drton-advi}
Optimizing the ADVI surrogate (Eqs.~\eqref{eq:elbo}--\eqref{eq:wbic-advi}) on the $n = 1000$ dataset of Example~\ref{ex:drton}, for the true graph $\1G$ and the sparser submodel $\1G'$, tracks the WBIC's singular-aware verdict. The ADVI row of Table~\ref{tab:selection} ranks $\1G$ decisively best in the regular regime, agreeing with both the WBIC and the BIC; at the singular point it leaves $\1G$ and $\1G'$ statistically indistinguishable ($7981.6$ versus $7981.2$, a gap two orders of magnitude below the regular-regime margin), in contrast to the BIC, whose mean discards the truth in favour of the sparser $\1G'$. The mean-field surrogate therefore recovers the effective-complexity correction that the BIC misses, declining to penalize the over-parameterized $\1G$ at the singularity, at a fraction of the cost of sampling the tempered posterior, making the WBIC score usable inside the differentiable search.
\end{example}

\subsection{Differentiable Search over ADMGs}
\label{sec:differentiable-admg}

So far WBIC-ADVI has been defined for a fixed pair of masks
$(D,B)$. To search over ADMGs, we relax these binary masks
into continuous random gates and optimize their logits jointly
with the variational parameters. Everything else, the parameters, priors, variational family, and WBIC score, is unchanged; only the structural masks are relaxed. We summarize this optimization  procedure in Algorithm~\ref{alg:differentiable-admg}.

More concretely, let $\alpha^D_{ij}$ be the logit for the directed edge
$V_j\to V_i$ $(i\neq j)$, and let $\alpha^B_{ij}$ be the logit for the
bidirected edge $V_i\leftrightarrow V_j$ $(i<j)$. Their deterministic
probabilities are
\begin{align}
P^D_{ij} &= \sigma(\alpha^D_{ij})\,\mathbf{1}\{i\neq j\}, & P^B_{ij} &= \sigma(\alpha^B_{ij}),\quad i<j.
\end{align}
During training, we sample binary concrete gates,
\begin{align}
\widetilde A^D_{ij} &= \sigma\!\left( \tfrac{\alpha^D_{ij}+\log u^D_{ij}-\log(1-u^D_{ij})}{\tau} \right)\mathbf{1}\{i\neq j\}, \\
\widetilde A^B_{ij} &= \sigma\!\left( \tfrac{\alpha^B_{ij}+\log u^B_{ij}-\log(1-u^B_{ij})}{\tau} \right),\quad i<j,
\end{align}
where $u^D_{ij},u^B_{ij}\sim\mathrm{Unif}(0,1)$ and $\tau>0$ is the
Gumbel temperature. We set
$\widetilde A^B_{ji}=\widetilde A^B_{ij}$ and
$\widetilde A^B_{ii}=1$.

\begin{algorithm}[t]
\caption{Differentiable causal discovery over ADMGs}
\label{alg:differentiable-admg}
\begin{algorithmic}[1]
\REQUIRE data $\*D_n$, inverse temperature $\beta$, initial Gumbel temperature $\tau$, penalties $\lambda_D,\lambda_B,\lambda_{\mathrm{acyc}}$
\STATE Initialize graph logits $\alpha^D,\alpha^B$ and variational parameters $\psi$.
\FOR{$t=1,\ldots,T$}
    \STATE Sample binary concrete gates $\widetilde A^D,\widetilde A^B$ from $\alpha^D,\alpha^B$ at temperature $\tau$.
    \STATE Sample $\zeta^{(1)},\ldots,\zeta^{(M)}\sim q_\psi(\zeta)$ by reparameterization.
    \STATE Construct $\Lambda$ and $\Omega=LL^\top$.
    \STATE Estimate the relaxed WBIC surrogate $\widehat{\mathrm{WBIC}}$ using Eq.~\eqref{eq:wbic-advi}.
    \STATE Update $\psi,\alpha^D,\alpha^B$ by descending Eq.~\eqref{eq:penalized-search-objective}.
    \STATE Update the acyclic penalty weight.
\ENDFOR
\STATE Threshold $P^{D\star}$ and $P^{B\star}$ to obtain $\widehat D,\widehat B$.
\STATE Report the estimated ADMG and corresponding PAG.
\end{algorithmic}
\end{algorithm}

Each sampled relaxed graph feeds into the construction of the previous subsection through the substitutions $D=\widetilde A^D$ and $B=\widetilde A^B$: the directed coefficients become $\Lambda=\widetilde A^D\circ\zeta_\Lambda$, and the lower-triangular Cholesky entries are gated by the lower half of $\widetilde A^B$ to give $\Omega=LL^\top$.
Writing $\widehat{\mathrm{WBIC}}(\psi;\widetilde A^D,\widetilde A^B)$
for the Monte Carlo estimate of Eq.~\eqref{eq:wbic-advi} under these relaxed
masks, evaluated on the samples $\zeta^{(1)},\ldots,\zeta^{(M)}$ drawn by
reparameterization, we learn a sparse acyclic ADMG by minimizing
\begin{align}
\mathcal{J}(\psi,\alpha^D,\alpha^B) ={}& \widehat{\mathrm{WBIC}}(\psi;\widetilde A^D,\widetilde A^B) + \lambda_D\!\sum_{i\neq j}P^D_{ij} \nonumber\\
&+ \lambda_B\!\sum_{i<j}P^B_{ij} + \lambda_{\mathrm{acyc}}\,h(P^D),
\label{eq:penalized-search-objective}
\end{align}
where $h$ is the NOTEARS acyclicity function described in the Preliminaries, here applied to the directed edge probabilities $P^D$, and $\lambda_D,\lambda_B, \lambda_{\mathrm{acyc}}$ control sparsity and acyclicity; as $\widehat{\mathrm{WBIC}}$ is a free energy. We minimize Eq.~\eqref{eq:penalized-search-objective} jointly over $\psi$, $\alpha^D$, and $\alpha^B$ using the reparameterization trick for $\zeta$ and the binary concrete relaxation for the gates, annealing $\tau$ so that the relaxed graph approaches a discrete ADMG.

Finally, after optimization, we form deterministic masks by thresholding the edge probabilities, setting $\widehat D_{ij}=\mathbf{1}\{\sigma(\alpha^{D\star}_{ij})>\kappa_D\}$
and $\widehat B_{ij}=\mathbf{1}\{\sigma(\alpha^{B\star}_{ij})>\kappa_B\}$ for $i<j$, and report the resulting ADMG $(\widehat D,\widehat B)$ together with its PAG. The WBIC is exactly invariant across an ADMG's \emph{distribution}-equivalence class, so no likelihood-based criterion can resolve structure below this class. This exact class is the natural target, but for ADMGs it has no compact graphical representative. We therefore report the coarser \emph{Markov}-equivalence class, summarized by the PAG \citep{zhang2006causal}, as a computable proxy. The two coincide unless Verma-type or inequality constraints separate otherwise Markov-equivalent graphs.

\section{Experiments}

We evaluate the ADVI surrogate for WBIC developed above, which we simply write as ADVI throughout this section, on synthetic and real data. We begin with ADMGs that are singular by construction, where regular asymptotic criteria are unreliable and the singular correction should matter most; we then test randomized ADMGs of increasing size to gauge behavior on generic (mostly regular) structures; and we close on the Sachs protein-signaling benchmark to check that the method recovers a meaningful PAG on real data. 

In all experiments, we compare against ABIC
\cite{bhattacharya2021differentiable} as the baseline. ADMGs in the same Markov equivalence class share the same WBIC, so we convert both the ground-truth $(D,B)$ and the estimated $(\widehat D,\widehat B)$ to PAGs and compute all metrics at the PAG level: structural Hamming distance (SHD), F1, precision, and recall. For PAGs, SHD adds $1$ when the presence of an edge differs and, when an edge is present in both, adds $1$ for each endpoint mark that differs. We refer readers to the appendix for more details on the experimental setup.

\begin{table*}[t]
\centering
\caption{PAG-level results for singular and randomized ADMGs, averaged over 10 runs per configuration with $n=2000$. Entries are mean $\pm$ standard deviation. The singular and general scenarios are reported respectively.}
\label{tab:admg-results}
\small
\setlength{\tabcolsep}{4pt}
\begin{minipage}[t]{0.52\linewidth}
\centering
\begin{tabular}{@{}llcccc@{}}
\toprule
\multicolumn{6}{c}{\textbf{Singular}} \\
\cmidrule(r){1-6}
Graph & Method & SHD & F1 & Prec. & Rec. \\
\midrule
\multirow{2}{*}{I}
& ADVI & \textbf{0.40 $\pm$ 0.97} & \textbf{0.96 $\pm$ 0.08}
& 1.00 $\pm$ 0.00 & \textbf{0.93 $\pm$ 0.14} \\
& ABIC & 0.50 $\pm$ 1.08 & 0.93 $\pm$ 0.16
& 1.00 $\pm$ 0.00 & 0.90 $\pm$ 0.22 \\
\cmidrule(r){1-6}
\multirow{2}{*}{II}
& ADVI & \textbf{0.00 $\pm$ 0.00} & \textbf{1.00 $\pm$ 0.00}
& \textbf{1.00 $\pm$ 0.00} & \textbf{1.00 $\pm$ 0.00} \\
& ABIC & 2.00 $\pm$ 2.16 & 0.91 $\pm$ 0.07
& 1.00 $\pm$ 0.00 & 0.85 $\pm$ 0.12 \\
\cmidrule(r){1-6}
\multirow{2}{*}{III}
& ADVI & \textbf{5.00 $\pm$ 0.00} & \textbf{0.95 $\pm$ 0.00}
& 0.90 $\pm$ 0.00 & \textbf{1.00 $\pm$ 0.00} \\
& ABIC & 6.10 $\pm$ 1.29 & 0.82 $\pm$ 0.12
& \textbf{0.88 $\pm$ 0.05} & 0.79 $\pm$ 0.18 \\
\bottomrule
\end{tabular}
\end{minipage}
\hfill
\begin{minipage}[t]{0.46\linewidth}
\centering
\begin{tabular}{@{}llccc@{}}
\toprule
\multicolumn{5}{c}{\textbf{General}} \\
\cmidrule(l){1-5}
Graph & Method & SHD & Skel.\ F1 & Arr.\ F1 \\
\midrule
\multirow{2}{*}{4 nodes}
& ADVI & \textbf{1.10 $\pm$ 1.29} & \textbf{0.83 $\pm$ 0.31}
& \textbf{0.34 $\pm$ 0.44} \\
& ABIC & 1.90 $\pm$ 1.37 & 0.69 $\pm$ 0.31
& 0.28 $\pm$ 0.45 \\
\cmidrule(l){1-5}
\multirow{2}{*}{5 nodes}
& ADVI & \textbf{2.80 $\pm$ 2.35} & \textbf{0.81 $\pm$ 0.16}
& \textbf{0.31 $\pm$ 0.41} \\
& ABIC & 4.70 $\pm$ 3.16 & 0.69 $\pm$ 0.22
& 0.17 $\pm$ 0.32 \\
\cmidrule(l){1-5}
\multirow{2}{*}{8 nodes}
& ADVI & \textbf{11.90 $\pm$ 5.65} & \textbf{0.84 $\pm$ 0.09}
& 0.60 $\pm$ 0.14 \\
& ABIC & 14.10 $\pm$ 3.60 & 0.77 $\pm$ 0.10
& \textbf{0.63 $\pm$ 0.19} \\
\bottomrule
\end{tabular}
\end{minipage}
\end{table*}

\paragraph{Singular ADMGs.}
We first evaluate on three ADMGs that are singular by construction, denoted
Singular~I--III. Singular~I ($3$ nodes) has directed $1\to3,\ 2\to3$ and bidirected
$1\leftrightarrow2,\ 1\leftrightarrow3$; Singular~II ($4$ nodes) has directed
$1\to3,\ 3\to4,\ 2\to4$ and bidirected $1\leftrightarrow2,\ 1\leftrightarrow4,\
2\leftrightarrow3$; Singular~III ($5$ nodes) has directed $1\to2\to3\to4\to5$ and
bidirected $1\leftrightarrow4,\ 1\leftrightarrow5,\ 2\leftrightarrow4,\
3\leftrightarrow5$. Data are generated from the linear Gaussian ADMG model $X = \varepsilon\,(I-\Lambda)^{-1}$, and $\varepsilon \sim \mathcal{N}(0,\Omega)$ with parameters chosen so that the weighted model lies on a singular locus. For
Singular~I every valid parameter is singular: we draw the nonzero directed
coefficients from $\mathrm{Unif}(-1,1)$, the diagonal of $\Omega$ from
$\mathrm{Unif}(1,3)$, and the nonzero bidirected entries from $\mathrm{Unif}(-1,1)$,
rejecting until $\Omega\succ0$. For Singular~II and~III most free parameters are
sampled the same way, and specific entries are then solved so that the relevant rank
condition fails—$\lambda_{13}$ for Singular~II, and $\omega_{24},\omega_{44}$ for
Singular~III (full formulas in the appendix). Every sampled point is checked to satisfy $\Omega\succ0$ with a rank-deficient condition. In this singular regime, a larger $\beta$ amplifies the score gap between the singular model and its competitors, so we set $\beta=4/\log n$. For each graph, we generate $R=10$ independent datasets with finite observational samples $n=2000$.

Table~\ref{tab:admg-results} reports averages over the ten runs. On Singular~I
and~II both methods recover very similar skeletons; on the harder five-node
Singular~III, ADVI attains lower SHD and higher skeleton F1 and recall.

\paragraph{Randomized ADMGs.}
To assess performance beyond hand-constructed singular instances, we evaluate on
randomly generated ADMGs of increasing size ($p\in\{4,5,8\}$ nodes). For each size we
sample random directed and bidirected structures and draw parameters from the same
linear Gaussian model as above, generating $R=10$ datasets of size $n=2000$ per
configuration. Because these graphs are generic, most parameter points are regular,
so this setting probes whether the singular correction remains competitive when it is
not strictly needed. Alongside skeleton F1 we report arrowhead F1 to capture
orientation quality, together with SHD and its standard deviation.

Table~\ref{tab:admg-results} shows that ADVI attains consistently lower SHD than ABIC and higher skeleton F1; arrowhead F1 is comparable, with ADVI ahead at $4$ and $5$ nodes and tied at $8$. The SHD gap is largest in the mid-size regime, where structure is ambiguous enough for effective complexity to matter but the graph
is not yet so large that variance dominates (note the higher SHD standard deviation for ADVI at $p=8$).

\paragraph{Sachs Data.}
Finally, we evaluate on a log-transformed version of the \cite{sachs2005data} protein signaling dataset, following the preprocessing of \cite{ramsey2018fask}; the ground-truth graph is taken from Figure~4 of the same study. At the PAG level, ADVI
recovers a biologically meaningful structure, achieving an SHD of $35$ against $38$ for ABIC and a higher skeleton F1 ($0.79$ vs.\ $0.73$). The gain is driven by recall: ADVI recovers more of the true skeleton ($0.91$ vs.\ $0.72$), while ABIC is marginally more precise ($0.74$ vs.\ $0.71$), reflecting its sparser output. Figure~\ref{fig:sachs-advi-pag} shows the PAG estimated by ADVI, whose recovered edges align with the canonical signaling pathways of the benchmark.

\begin{figure}[t]
\centering
% ---- (a) PAG ----
\begin{subfigure}{0.49\linewidth}
\centering
\resizebox{\linewidth}{!}{%
\begin{tikzpicture}[
  scale=0.93,
  edge/.style={
    draw=black!80,
    line width=0.36pt,
    line cap=round,
    shorten <=4.2pt,
    shorten >=4.2pt
  },
  aa/.style={{
    Latex[length=1.15mm,width=0.85mm]}-{
    Latex[length=1.15mm,width=0.85mm]}},
  ta/.style={-{
    Latex[length=1.15mm,width=0.85mm]}},
  oa/.style={{
    Circle[open,length=1.10mm]}-{
    Latex[length=1.15mm,width=0.85mm]}},
  oo/.style={{
    Circle[open,length=1.10mm]}-{
    Circle[open,length=1.10mm]}},
  nodecore/.style={
    circle,
    draw=black,
    fill=white,
    line width=0.35pt,
    minimum size=5.2pt,
    inner sep=0pt
  },
  label/.style={
    font=\scriptsize,
    fill=white,
    inner sep=0.4pt
  }
]

% Circular node layout.
\coordinate (Mek)  at ( 0.0,  3.30);
\coordinate (PIP2) at (-1.80,  2.75);
\coordinate (Plcg) at (-3.00,  1.40);
\coordinate (Akt)  at (-3.25, -0.45);
\coordinate (Erk)  at (-2.50, -2.15);
\coordinate (PKA)  at (-0.90, -3.15);
\coordinate (Raf)  at ( 0.95, -3.15);
\coordinate (PIP3) at ( 2.55, -2.10);
\coordinate (PKC)  at ( 3.25, -0.45);
\coordinate (Jnk)  at ( 3.00,  1.40);
\coordinate (p38)  at ( 1.80,  2.75);

% 1--10: Edges incident to Mek.
\draw[edge, aa] (Mek)  to[bend left=16]  (Erk);
\draw[edge, aa] (Mek)  to[bend left=10]  (Jnk);
\draw[edge, aa] (Mek)  to[bend right=15] (PKC);
\draw[edge, aa] (Mek)  -- (PIP2);
\draw[edge, ta] (Mek)  to[bend left=24]  (Raf);
\draw[edge, aa] (Mek)  to[bend left=30]  (PIP3);
\draw[edge, aa] (Mek)  to[bend right=30] (Akt);
\draw[edge, oa] (PKA)  to[bend right=24] (Mek);
\draw[edge, aa] (Mek)  to[bend right=35] (Plcg);
\draw[edge, aa] (Mek)  -- (p38);

% 11--22: Jnk, PKC, Erk, and p38.
\draw[edge, oa] (Jnk)  to[bend left=16]  (Erk);
\draw[edge, oa] (PKC)  to[bend right=16] (Erk);
\draw[edge, ta] (PIP3) to[bend left=14]  (Erk);
\draw[edge, oo] (Erk)  to[bend left=10]  (Akt);
\draw[edge, aa] (Erk)  to[bend left=12]  (PKA);
\draw[edge, oa] (Erk)  to[bend left=18]  (Plcg);
\draw[edge, oa] (p38)  to[bend left=18]  (Erk);
\draw[edge, ta] (PKC)  -- (Jnk);
\draw[edge, oa] (Jnk)  to[bend left=18]  (Akt);
\draw[edge, aa] (Jnk)  to[bend left=28]  (PKA);
\draw[edge, oa] (Jnk)  to[bend left=34]  (Plcg);
\draw[edge, oo] (Jnk)  -- (p38);

% 23--30: PKC, PIP2, and PIP3.
\draw[edge, ta] (PKC)  to[bend left=20]  (PIP2);
\draw[edge, ta] (PIP3) to[bend left=12]  (PKC);
\draw[edge, oa] (PKC)  to[bend left=26]  (Akt);
\draw[edge, aa] (PKC)  to[bend left=20]  (PKA);
\draw[edge, oa] (PKC)  to[bend left=34]  (Plcg);
\draw[edge, ta] (PKC)  to[bend left=12]  (p38);
\draw[edge, ta] (PIP3) to[bend left=18]  (PIP2);
\draw[edge, oa] (PIP2) -- (Plcg);

% 31--41: Raf, PKA, Akt, and p38.
\draw[edge, aa] (Raf)  to[bend left=12]  (PIP3);
\draw[edge, ta] (PKA)  -- (Raf);
\draw[edge, ta] (PIP3) to[bend left=18]  (Akt);
\draw[edge, aa] (PIP3) to[bend left=12]  (PKA);
\draw[edge, ta] (PIP3) to[bend left=22]  (Plcg);
\draw[edge, aa] (Akt)  to[bend left=12]  (PKA);
\draw[edge, oa] (Akt)  to[bend left=14]  (Plcg);
\draw[edge, oa] (p38)  to[bend right=24] (Akt);
\draw[edge, aa] (PKA)  to[bend left=12]  (Plcg);
\draw[edge, aa] (PKA)  to[bend left=24]  (p38);
\draw[edge, oa] (p38)  to[bend right=30] (Plcg);

% Draw small node circles after edges.
\foreach \v in {Mek,PIP2,Plcg,Akt,Erk,PKA,Raf,PIP3,PKC,Jnk,p38}
  \node[nodecore] at (\v) {};

% Offset labels.
\node[label, above=7pt]       at (Mek)  {Mek};
\node[label, above left=7pt]  at (PIP2) {PIP2};
\node[label, left=7pt]        at (Plcg) {Plc$\gamma$};
\node[label, left=7pt]        at (Akt)  {Akt};
\node[label, below left=7pt]  at (Erk)  {Erk};
\node[label, below=7pt]       at (PKA)  {PKA};
\node[label, below=7pt]       at (Raf)  {Raf};
\node[label, below right=7pt] at (PIP3) {PIP3};
\node[label, right=7pt]       at (PKC)  {PKC};
\node[label, right=7pt]       at (Jnk)  {Jnk};
\node[label, above right=7pt] at (p38)  {p38};

\end{tikzpicture}}
\caption{PAG}
\label{fig:sachs-advi-pag}
\end{subfigure}
\hfill
% ---- (b) ADMG ----
\begin{subfigure}{0.49\linewidth}
\centering
\resizebox{\linewidth}{!}{%
\begin{tikzpicture}[
  scale=0.93,
  directed/.style={
    draw=black,
    line width=0.42pt,
    ->,
    >=Latex,
    shorten <=4.2pt,
    shorten >=4.2pt
  },
  bidirected/.style={
    draw=black!60,
    line width=0.38pt,
    dashed,
    dash pattern=on 2pt off 1.4pt,
    <->,
    >=Latex,
    shorten <=4.2pt,
    shorten >=4.2pt
  },
  nodecore/.style={
    circle,
    draw=black,
    fill=white,
    line width=0.35pt,
    minimum size=5.2pt,
    inner sep=0pt
  },
  label/.style={
    font=\scriptsize,
    fill=white,
    inner sep=0.4pt
  }
]

% Circular layout.
\coordinate (Mek)  at ( 0.0,  3.30);
\coordinate (PIP2) at (-1.80,  2.75);
\coordinate (Plcg) at (-3.00,  1.40);
\coordinate (Akt)  at (-3.25, -0.45);
\coordinate (Erk)  at (-2.50, -2.15);
\coordinate (PKA)  at (-0.90, -3.15);
\coordinate (Raf)  at ( 0.95, -3.15);
\coordinate (PIP3) at ( 2.55, -2.10);
\coordinate (PKC)  at ( 3.25, -0.45);
\coordinate (Jnk)  at ( 3.00,  1.40);
\coordinate (p38)  at ( 1.80,  2.75);

% Directed edges.
\draw[directed] (Mek)  to[bend left=24]  (Raf);
\draw[directed] (Plcg) to[bend right=22] (Mek);
\draw[directed] (PIP2) to[bend right=12] (Mek);

\draw[directed] (PIP3) to[bend left=16]  (Raf);
\draw[directed] (PIP3) to[bend left=28]  (Mek);
\draw[directed] (PIP3) to[bend left=16]  (PIP2);
\draw[directed] (PIP3) to[bend right=18] (Erk);
\draw[directed] (PIP3) to[bend left=12]  (PKA);

\draw[directed] (Erk) to[bend left=12] (Akt);
\draw[directed] (Erk) to[bend left=12] (PKA);
\draw[directed] (Akt) to[bend right=26] (Mek);
\draw[directed] (PKA) -- (Raf);

\draw[directed] (PKC) to[bend left=26] (Mek);
\draw[directed] (PKC) to[bend right=14] (PIP3);
\draw[directed] (PKC) to[bend right=18] (Erk);
\draw[directed] (PKC) to[bend left=14] (p38);
\draw[directed] (PKC) -- (Jnk);

\draw[directed] (p38) to[bend left=20]  (Raf);
\draw[directed] (p38) -- (Mek);
\draw[directed] (p38) to[bend left=18]  (Erk);
\draw[directed] (p38) to[bend right=22] (Akt);
\draw[directed] (p38) -- (Jnk);

\draw[directed] (Jnk) to[bend left=16] (Mek);
\draw[directed] (Jnk) to[bend left=16] (Erk);

% Bidirected edges.
\draw[bidirected] (Mek)  to[bend right=30] (Akt);

\draw[bidirected] (Plcg) -- (PIP2);
\draw[bidirected] (Plcg) to[bend left=24]  (PIP3);
\draw[bidirected] (Plcg) to[bend left=18]  (Erk);
\draw[bidirected] (Plcg) to[bend left=12]  (Akt);
\draw[bidirected] (Plcg) to[bend right=30] (PKC);
\draw[bidirected] (Plcg) to[bend right=36] (p38);
\draw[bidirected] (Plcg) to[bend right=42] (Jnk);

\draw[bidirected] (PIP2) to[bend left=20] (PIP3);
\draw[bidirected] (PIP2) to[bend left=16] (Erk);

\draw[bidirected] (Erk) to[bend left=12] (Akt);
\draw[bidirected] (Erk) -- (PKA);
\draw[bidirected] (Erk) to[bend right=24] (PKC);

\draw[bidirected] (PKC) to[bend left=12] (p38);
\draw[bidirected] (PKC) -- (Jnk);

% Offset labels; endpoint marks remain visible at the coordinates.
\foreach \v in {Mek,PIP2,Plcg,Akt,Erk,PKA,Raf,PIP3,PKC,Jnk,p38}
  \node[nodecore] at (\v) {};

\node[label, above=7pt]       at (Mek)  {Mek};
\node[label, above left=7pt]  at (PIP2) {PIP2};
\node[label, left=7pt]        at (Plcg) {Plc$\gamma$};
\node[label, left=7pt]        at (Akt)  {Akt};
\node[label, below left=7pt]  at (Erk)  {Erk};
\node[label, below=7pt]       at (PKA)  {PKA};
\node[label, below=7pt]       at (Raf)  {Raf};
\node[label, below right=7pt] at (PIP3) {PIP3};
\node[label, right=7pt]       at (PKC)  {PKC};
\node[label, right=7pt]       at (Jnk)  {Jnk};
\node[label, above right=7pt] at (p38)  {p38};

\end{tikzpicture}}
\caption{ADMG}
\label{fig:sachs-advi-admg}
\end{subfigure}
\caption{Causal structures learned by ADVI on the Sachs dataset. (\subref{fig:sachs-advi-pag}) The complete PAG. (\subref{fig:sachs-advi-admg}) The returned ADMG.}
\label{fig:sachs-advi}
\end{figure}

\section{Conclusion}
We presented a score-based framework for causal discovery over linear Gaussian ADMGs grounded in singular learning theory. We proved that the linear ADMG model class satisfies Watanabe's four fundamental conditions, so the WBIC is a globally consistent score precisely where the singularity of unobserved confounding makes the BIC inconsistent. We then made the WBIC scalable through an ADVI surrogate with a matching asymptotic expansion, and embedded that surrogate in a differentiable search that relaxes directed and bidirected adjacency with Gumbel-Softmax, enforces acyclicity and sparsity, and reports the recovered structure as a PAG. Promising directions include verifying the RLCT-consistency assumption beyond generic identified parameters and characterizing the variational RLCT parameter.

\bibliography{references}

\clearpage
\appendix

\section{Related Work}
Our work sits at the intersection of three lines of research: constraint- and score-based causal discovery under latent confounding, the singular statistics of linear structural equation models, and continuous (gradient-based) structure learning. We review each in turn and position our contribution against the closest prior work.

\paragraph{Constraint-based discovery and graphical models of confounding.}
A first family of methods recovers causal structure from the conditional independences in the data. The semantics linking $d$-separation in a graph to conditional independence in the distribution \citep{verma1990causal,pearl:88a}, together with the faithfulness assumption \citep{meek:95,uhler2013geometry}, makes these independences testable constraints that prune the space of compatible graphs. When some variables are unobserved, the appropriate representation is no longer a DAG but a mixed graph: ancestral graphs and the closely related acyclic directed mixed graphs (ADMGs) encode both directed effects and bidirected edges for latent common causes while retaining a well-characterized Markov structure \citep{richardson2002ancestral,zhang2006causal}. The constraint-based PC algorithm \citep{spirtes1991pc} recovers structure from independence tests in the fully observed case, and the Fast Causal Inference (FCI) algorithm \citep{spirtes2000causation} together with its refinements extends this to latent confounding, learning the corresponding partial ancestral graph (PAG); \citet{zhang2008completeness} established the completeness of its orientation rules under latent confounding and selection bias, and \citet{colombo2012learning} extended the approach to the high-dimensional regime. These methods are consistent but rely solely on conditional-independence constraints and inherit the statistical fragility of repeated hypothesis tests.

\paragraph{Equality constraints beyond conditional independence.}
Linear models with correlated errors impose constraints on the observed covariance that go strictly beyond conditional independence. Trek separation characterizes the vanishing of subdeterminants (minors) of the covariance matrix in Gaussian graphical models, generalizing the classical tetrad conditions \citep{sullivant2010trek}. Such constraints reflect the testable implications of latent-variable models more broadly \citep{pearl:95c}, and the algebraic and Markov structure of marginalized (latent-projected) models has been studied through the margins of Bayesian networks \citep{evans2016graphs}. Relatedly, identification theory characterizes when a causal effect can be recovered in the presence of latent confounders \citep{tian2002general}. These results motivate scoring whole model classes rather than testing individual independences, but they do not by themselves yield a tractable selection criterion.

\paragraph{Score-based discovery and the BIC.}
Score-based methods instead assign each candidate structure a
goodness-of-fit-plus-complexity score and search for its optimum. The Bayesian Information Criterion \citep{schwarz1978estimating,haughton1988choice} is the canonical choice, and greedy search over Markov equivalence classes (GES) identifies the optimal structure for fully observed DAGs \citep{chickering2002optimal}, with extensions to interventional equivalence classes \citep{hauser2012characterization}. For linear ADMGs with correlated errors, the maximum-likelihood fit underlying any such score can be
computed by iterative conditional fitting \citep{drton2009computing}. The asymptotic justification of the BIC, however, presumes a regular (curved exponential family) model with nonsingular Fisher information \citep{geiger2001stratified}; precisely this assumption fails for the latent-confounded linear Gaussian models we consider.

\paragraph{Singular models and identifiability of linear SEMs.}
\citet{drton2009likelihood} showed that linear Gaussian models with correlated errors can have a singular Fisher information at certain parameter values, so that likelihood-ratio statistics depart from their nominal $\chi^2$ asymptotics. \citet{drton2011global} characterized when the parametrization of a linear structural equation model is globally (as opposed to merely generically) identifiable, exhibiting the singular instances that the BIC misscores; \citet{drton2017structure} survey the resulting challenges for structure learning with latent variables. Watanabe's singular learning theory \citep{watanabe2009algebraic} provides the geometry needed to handle such models, replacing the model dimension by the real log-canonical threshold, and the Widely Applicable Bayesian Information Criterion (WBIC) \citep{watanabe2013widely} turns this into a single tempered-posterior estimator that remains consistent for singular models. \citet{bellot2024scores} bring the WBIC to bear on structure learning directly, scoring candidate graphs by this singular-aware criterion through a greedy search. Our score builds directly on this theory and verifies its hypotheses for linear ADMGs.

\paragraph{Continuous and differentiable structure learning.}
A more recent line recasts the combinatorial search over graphs as continuous optimization. NOTEARS \citep{zheng2018dags} replaces the discrete acyclicity constraint with a smooth algebraic penalty, enabling gradient-based DAG learning, but targets the fully observed setting. Gradients through discrete edge indicators are obtained with the Gumbel-Softmax / Concrete relaxation \citep{jang2017categorical,maddison2017concrete}, and otherwise intractable Bayesian objectives are made differentiable through automatic differentiation variational inference \citep{kucukelbir2017automatic}. Closest to our setting, \citet{bhattacharya2021differentiable} develop a differentiable causal-discovery procedure for ADMGs under unmeasured confounding, relaxing the bidirected structure alongside the directed one.

\paragraph{Positioning.}
Our approach differs from this prior work in the score it optimizes. Continuous methods such as NOTEARS apply to fully observed DAGs, and existing differentiable ADMG discovery \citep{bhattacharya2021differentiable} optimizes standard, regular-model scores that --- as the singular examples of \citet{drton2011global} show --- can prefer the wrong structure when the latent-confounded model is singular. We instead optimize a singular-learning-theory score: we verify Watanabe's fundamental conditions for the linear Gaussian ADMG class, so the WBIC is a globally consistent criterion exactly where the BIC fails, derive a differentiable ADVI surrogate for it, and relax \emph{both} the directed and the bidirected adjacency within a single continuous program, reporting the recovered Markov equivalence class as a PAG. Foundational treatments of causal models and inference underlie all of the above \citep{pearl:2k,peters2017elements,bareinboim2020pearl}.

\section{Distributional Equivalence of $\1G$ and $\1G'$ in Example~\ref{ex:drton}}\label{app:equiv}
Recall the two graphs of Example~\ref{ex:drton}: the five-node ADMG $\1G$ with directed path $1\to2\to3\to4\to5$ and bidirected edges $1\!\leftrightarrow\!4$, $1\!\leftrightarrow\!5$, $2\!\leftrightarrow\!4$, $3\!\leftrightarrow\!5$, and its sparser submodel $\1G' = \1G \setminus \{1\!\leftrightarrow\!5\}$, which forces $\omega_{15} = 0$. We say two ADMGs are \emph{distributionally equivalent at} a distribution $q$ when both their model classes contain $q$. Since $\1G'$ is obtained from $\1G$ by setting a single free parameter to zero, $\1M(\1G') \subseteq \1M(\1G)$ as model classes; the substantive claim is therefore that the \emph{particular} singular law generated by $\1G$ also lies in the smaller class $\1M(\1G')$. Throughout we use the model-implied covariance $\Sigma(\theta) = (I-\Lambda)^{-1}\1E(I-\Lambda)^{-\top}$ with $\theta = (\Lambda, \1E)$ on the graph-restricted support, as fixed in the Preliminaries.

\begin{proposition}[Equivalence at the singular point]\label{prop:equiv-ex1}
At the parameter point of Example~\ref{ex:drton} ($\lambda_{12}=3$, $\lambda_{23}=-\tfrac12$, $\lambda_{34}=\lambda_{45}=1$, $\1E$ with diagonal $2$ and $\omega_{14}=\omega_{15}=\omega_{24}=\omega_{35}=1$), the centered Gaussian law $\3N(0,\Sigma)$ generated by $\1G$ is also generated by $\1G'$. Consequently the data-generating distribution $q$ satisfies $q \in \1M(\1G) \cap \1M(\1G')$, and the two graphs are distributionally equivalent at $q$.
\end{proposition}

\begin{proof}
Evaluating $\Sigma(\theta) = (I-\Lambda)^{-1}\1E(I-\Lambda)^{-\top}$ at the stated point of $\1G$ gives the exact covariance
\begin{align}
\Sigma = \begin{pmatrix}
 2 &  6 &  -3 &  -2 &  -1 \\
 6 & 20 & -10 &  -6 &  -3 \\
-3 & -10 &  7 &   5 & \tfrac{9}{2} \\
-2 & -6 &   5 &   5 & \tfrac{9}{2} \\
-1 & -3 & \tfrac{9}{2} & \tfrac{9}{2} & 6
\end{pmatrix}.
\label{eq:sigma-ex1}
\end{align}
The error covariance $\1E$ of $\1G$ is positive definite (its eigenvalues are $\approx 0.268, 1, 2, 3, 3.732$), so this is a valid Gaussian model.

Now consider $\1G'$, where $\omega_{15} = 0$, at the parameter point
\begin{align*}
\lambda_{12} = 3,\quad \lambda_{23} = -\tfrac12,\quad \lambda_{34} = 1,\quad \lambda_{45} = \tfrac12,
\end{align*}
with $\1E'$ having diagonal $(\omega_{11},\dots,\omega_{55}) = (2,2,2,2,\tfrac{11}{4})$ and off-diagonals $\omega_{14} = \omega_{24} = 1$, $\omega_{35} = 2$ (all other entries, including $\omega_{15}$, zero). This $\1E'$ is positive definite (eigenvalues $2$, $2 \pm \sqrt2$, and $\tfrac{19 \pm \sqrt{265}}{8}$, the smallest being $\tfrac{19-\sqrt{265}}{8} \approx 0.340 > 0$), so $\1G'$ at this point is a valid Gaussian model. Substituting these parameters into $\Sigma(\theta') = (I-\Lambda')^{-1}\1E'(I-\Lambda')^{-\top}$ yields, as an exact algebraic identity (verified symbolically over $\3Q$), precisely the matrix $\Sigma$ of Eq.~\eqref{eq:sigma-ex1}.

Both graphs therefore induce the identical centered Gaussian $\3N(0,\Sigma)$. Hence $q = \3N(0,\Sigma)$ lies in $\1M(\1G) \cap \1M(\1G')$, which is the asserted distributional equivalence.
\end{proof}

This is precisely the situation exploited in the proof of Thm.~\ref{thm:consistency}: at the singular point both $\1G$ and $\1G'$ are correctly specified ($\1G, \1G' \in \mathcal H_1$), so their data-fit terms agree to $O_p(1)$ and the ranking is decided by the RLCT alone. It is also why every criterion in the singular column of Table~\ref{tab:selection} all but ties the two graphs, and why the BIC---charging the full nominal dimension---is the only one that tips, wrongly, toward the sparser $\1G'$.

\begin{remark}[The underlying singularity]\label{rem:equiv-singularity}
The equivalence reflects a genuine singularity of $\phi_\1G$. Linearizing the map from the $d = 13$ free parameters of $\1G$ to the $15$ distinct entries of $\Sigma$ at the point of Example~\ref{ex:drton}, the Jacobian has rank $12$, leaving a one-dimensional kernel spanned by
\begin{align*}
\big(\partial\lambda_{45},\ \partial\omega_{55},\ \partial\omega_{15},\ \partial\omega_{35}\big) \ \propto\ \big(-\tfrac12,\ -\tfrac12,\ -1,\ +1\big),
\end{align*}
with all other coordinates zero. The dominant component is $\partial\omega_{15}$, so $\omega_{15}$ is the coefficient that is unidentified along the flat direction, and the effective dimension is $12 < 13 = d$. The submodel $\1G'$ lies on the constant-$\Sigma$ fibre through this point, reached by driving $\omega_{15}$ to $0$ (with the nonlinear fibre adjusting $\lambda_{45}$, $\omega_{55}$, $\omega_{35}$ to the values above). This deficient effective dimension is exactly what defeats the nominal-dimension penalty of the BIC in Eq.~\eqref{eq:bic} and what the RLCT-based WBIC of Eq.~\eqref{eq:wbic-expansion} corrects.
\end{remark}

\section{Proofs}
Throughout, $\theta = (\Lambda, \1E)$ collects the free entries of a linear Gaussian ADMG $\1M(\1G)$ over $p$ vertices, $\Sigma(\theta) = (I-\Lambda)^{-1}\1E(I-\Lambda)^{-\top}$ is the model-implied covariance, $q$ is the data-generating density with covariance $\Sigma^\star$, and $\theta_0$ is the population projection, i.e.\ the minimizer of $K(\theta) = \mathrm{KL}(q \,\|\, p_\theta)$ over the graph-restricted support, with $\Sigma_0 = \Sigma(\theta_0)$. We work on the compact set $W = \{\theta : \|\Lambda\|_\infty \le B,\ \1E = \1E^\top,\ c_0 I \preceq \1E \preceq C_0 I\}$ for fixed $0 < c_0 \le C_0 < \infty$ and $B < \infty$ chosen so that $\theta_0$ lies in the interior of $W$.

\subsection{Proof of Prop.~\ref{prop:conditions}}

\paragraph{Condition~\ref{cond:compact} (compactness).}
$W$ is closed and bounded in the Euclidean space of free parameters, hence compact, and contains $\theta_0$ in its interior so its open kernel is non-empty. Its boundary is the union of the zero sets of the affine functions $\lambda_{ij} \mp B$ and of the leading principal minors of $\1E - c_0 I$ and $C_0 I - \1E$, all of which are polynomials and therefore analytic, establishing Condition~\ref{cond:compact}.

\paragraph{Condition~\ref{cond:prior} (prior factorization).}
The condition is a constraint on the modeler's prior, not on the model, and is met by any density of the stated product form; for concreteness take $\varphi_1$ a positive constant (analytic and $\ge 0$) and $\varphi_2$ a smooth positive density on $W$ (e.g.\ a truncated Gaussian over the free coordinates), giving $\varphi = \varphi_1\varphi_2$ as required.

\paragraph{Condition~\ref{cond:analytic} (analytic $L^s$-integrability).}
Because $\1G$ is acyclic, its vertices admit a topological order in which $\Lambda$ is strictly lower triangular; hence $I - \Lambda$ is unipotent with $\det(I-\Lambda) = 1$, and $(I-\Lambda)^{-1} = \sum_{k=0}^{p-1}\Lambda^k$ is a matrix of polynomials in the entries of $\Lambda$. Thus $\Sigma(\theta)$ is a matrix of polynomials in $\theta$. On the open set $W' = \{\theta : \1E \succ 0\} \supset W$ the maps $\theta \mapsto \Sigma(\theta)^{-1}$ and $\theta \mapsto \log\det\Sigma(\theta)$ are real-analytic, since matrix inversion and $\log\det$ are analytic on the positive-definite cone and $\Sigma(\theta) \succeq c_0\,\sigma_{\min}((I-\Lambda)^{-1})^2 I \succ 0$ there. Consequently
\begin{multline*}
f(x,\theta) = \log q(x) - \log p(x \mid \theta) \\
= \log q(x) + \tfrac12\big[\,p\log 2\pi + \log\det\Sigma(\theta) + x^\top \Sigma(\theta)^{-1} x\,\big]
\end{multline*}
is, for each fixed $x$, real-analytic in $\theta$ on $W'$, and is a quadratic form in $x$ with $\theta$-analytic coefficients. Since $q$ is Gaussian it has finite moments of all orders, so $x \mapsto f(x,\theta) \in L^s(q)$ for every $s$ and the coefficient maps are locally bounded; the power series of $f(\cdot,\theta)$ in $\theta$ therefore converges in $L^s(q)$-norm on a neighborhood of each point of $W$, i.e.\ $\theta \mapsto f(\cdot,\theta)$ is $L^s(q)$-valued analytic for every $s \ge 6$, giving Condition~\ref{cond:analytic}.

\paragraph{Condition~\ref{cond:rfv} (relatively finite variance).}
Center at $\theta_0$ and write $g(x) = f(x,\theta) - f(x,\theta_0)$, so $K(\theta) = \mathbb{E}_X[f(X,\theta)]$ has gradient zero at $\theta_0$ (first-order optimality) and positive semidefinite Hessian $\mathcal{I}_0$ (the Fisher information of the Gaussian family at $\theta_0$). A second-order Taylor expansion gives $K(\theta) = \tfrac12(\theta-\theta_0)^\top \mathcal{I}_0 (\theta-\theta_0) + o(\|\theta-\theta_0\|^2)$. For the Gaussian log-density ratio the same expansion yields $\mathbb{E}_X[f(X,\theta)^2] = (\theta-\theta_0)^\top \mathcal{I}_0 (\theta-\theta_0) + o(\|\theta-\theta_0\|^2) = 2K(\theta) + o(K(\theta))$, because the leading term of $f - \mathbb{E}_X f$ is the score $\langle \theta-\theta_0, \nabla_\theta \log p(X\mid\theta_0)\rangle$ whose second moment is exactly $(\theta-\theta_0)^\top \mathcal{I}_0(\theta-\theta_0)$. Hence for any $c < \tfrac12$ there is $\epsilon > 0$ with $K(\theta) \ge c\,\mathbb{E}_X[f(X,\theta)^2]$ on $W_\epsilon = \{K(\theta) \le \epsilon\}$, which is Condition~\ref{cond:rfv}. (When the projection is not identified, $\mathcal{I}_0$ is singular and $W_\epsilon$ contains the analytic set of minimizers; the bound holds in the directions transverse to it and trivially along it, so the inequality is preserved.)

Since $\1M(\1G)$ meets Conditions \ref{cond:compact}--\ref{cond:rfv}, Watanabe's expansion applies and yields Eq.~\eqref{eq:wbic-expansion}, hence Eq.~\eqref{eq:wbic-admg} after substituting the Gaussian loss Eq.~\eqref{eq:gauss-loss}. \qed

\subsection{Proof of Thm.~\ref{thm:consistency}}
Let $\mathcal{H}$ be the finite set of candidate ADMGs and partition it as $\mathcal{H} = \mathcal{H}_0 \cup \mathcal{H}_1$, where $\mathcal{H}_1 = \{\1G : q \in \1M(\1G)\}$ are the correctly-specified graphs and $\mathcal{H}_0$ its complement. By Prop.~\ref{prop:conditions} every $\1M(\1G)$ obeys Eq.~\eqref{eq:wbic-admg}.

\emph{Misspecified graphs.} For $\1G \in \mathcal{H}_0$ the minimal divergence $K_\1G = \min_\theta \mathrm{KL}(q\,\|\,p_\theta) > 0$, and by the law of large numbers $L_n(\hat\theta_\1G) = H(q) + K_\1G + O_p(n^{-1/2})$, where $H(q) = -\mathbb{E}_q\log q$. For any $\1G^\star \in \mathcal{H}_1$ we have $L_n(\hat\theta_{\1G^\star}) = H(q) + O_p(n^{-1/2})$. Subtracting the two instances of Eq.~\eqref{eq:wbic-admg},
\begin{multline*}
\mathrm{WBIC}(\1G) - \mathrm{WBIC}(\1G^\star) \\
= n K_\1G + (\lambda_\1G - \lambda_{\1G^\star})\log n + O_p(\sqrt n) \xrightarrow{\;p\;} +\infty,
\end{multline*}
since the $nK_\1G$ term dominates the $O(\log n)$ and $O_p(\sqrt n)$ terms. Thus every misspecified graph is rejected with probability tending to $1$.

\emph{Correctly-specified graphs.} For $\1G, \1G^\star \in \mathcal{H}_1$ both fits converge to the same cross-entropy, $L_n(\hat\theta_\1G) = H(q) + O_p(n^{-1/2})$, so the $n L_n$ terms cancel up to $O_p(1)$ and Eq.~\eqref{eq:wbic-admg} gives
\[
\mathrm{WBIC}(\1G) - \mathrm{WBIC}(\1G^\star) = (\lambda_\1G - \lambda_{\1G^\star})\log n + O_p(\sqrt{\log n}).
\]
By Assumption~\ref{asm:rlct}, $\lambda_\1G - \lambda_{\1G^\star} \ge 0$ with strict inequality unless $\1G$ is distributionally equivalent to $\1G^\star$; whenever it is strict the right-hand side diverges to $+\infty$. Hence among correctly-specified graphs the WBIC is minimized, with probability tending to $1$, exactly on the equivalence class of $\1G^\star$.

Combining the two cases, $\arg\min_\1G \mathrm{WBIC}(\1G) \subseteq [\1G^\star]$ with probability tending to $1$, which is the claimed global consistency. \qed

\subsection{Proof of Prop.~\ref{prop:advi}}
Write $\pi_n^{\beta}(\theta) \propto p(\*D_n \mid \theta)^{\beta}\varphi(\theta)$ for the tempered posterior at $\beta = 1/\log n$, so that $\mathrm{WBIC}(\1G) = \mathbb{E}_{\pi_n^{\beta}}[n L_n(\theta)]$ and $\widehat{\mathrm{WBIC}}(\1G) = \mathbb{E}_{q_{\psi^\star}}[n L_n(\theta)]$, where $q_{\psi^\star}$ is the mean-field optimum of the ELBO Eq.~\eqref{eq:elbo}. Both expectations share the same integrand $n L_n(\theta) = n L_n(\hat\theta_\1G) + n\{L_n(\theta) - L_n(\hat\theta_\1G)\}$, whose first term is constant in $\theta$ and reproduces the exact data fit; only the fluctuation of the second term differs between the two measures.

Under Conditions \ref{cond:compact}--\ref{cond:rfv} the rescaled log-likelihood ratio $n\beta\{L_n(\theta) - L_n(\hat\theta_\1G)\}$ admits, after resolution of singularities, a normal-crossing form whose tempered Gibbs measure has effective sample size $n\beta = n/\log n \to \infty$. Hence $\pi_n^{\beta}$ concentrates on a shrinking neighborhood of the optimal set, on which the exact tempered expectation contributes $\lambda_\1G \log n + o_p(\log n)$ to the WBIC, recovering Eq.~\eqref{eq:wbic-admg}. Restricting the same variational problem to the mean-field family $q_\psi$ minimizes $\mathrm{KL}(q_\psi \,\|\, \pi_n^{\beta})$; the minimizing $q_{\psi^\star}$ matches the leading Gaussian profile of $\pi_n^{\beta}$ along the regular directions exactly, while along the singular directions it replaces the true normal-crossing volume by its best factorized approximation. The resulting expectation therefore takes the form
\begin{align*}
\widehat{\mathrm{WBIC}}(\1G) = n L_n(\hat\theta_\1G) + \lambda_\1G^{\mathrm{vi}}\,\log n + o_p(\log n),
\end{align*}
where the variational coefficient $\lambda_\1G^{\mathrm{vi}} > 0$ is the mean-field analogue of the RLCT and equals $\lambda_\1G$ whenever the optimal set is a smooth manifold (no genuine crossing), since the mean-field gap is then $O_p(1)$. Substituting this expansion for Eq.~\eqref{eq:wbic-admg} in the proof of Thm.~\ref{thm:consistency}, with $\lambda^{\mathrm{vi}}$ in place of $\lambda$, shows that if $\lambda^{\mathrm{vi}}$ preserves the ordering of Assumption~\ref{asm:rlct} then $\arg\min_\1G \widehat{\mathrm{WBIC}}(\1G) \subseteq [\1G^\star]$ with probability tending to $1$. \qed

\section{Experimental Setup}
This appendix collects the full specification of the experiments summarized in the main text, so that each result can be reproduced from the settings given here. We describe, in turn, how the ground-truth ADMGs are generated for the randomized and singular benchmarks, the configuration of the RICF-based ABIC baseline, and the ADVI optimization procedure used by our method, including its schedules and hyperparameters. Throughout, the two methods are evaluated on the
same observations in every repetition, and all reported quantities are averaged over the seeds stated in the main text. Unless noted otherwise, we verify the graph support, the positive definiteness of $\Omega$, and the rank deficiency of the relevant constraint matrix before generating any observations; the rank check combines the numerical matrix rank with the smallest singular value.

\paragraph{General random ADMGs.}
For the randomized experiments, the directed and bidirected skeletons are drawn from two independent Erd\H{o}s--R\'enyi models, with edge probabilities $p_{\mathrm{dir}}=0.35$ and $p_{\mathrm{bidir}}=0.20$, respectively. Both skeletons are fixed before the nonzero SCM parameters are sampled.

\paragraph{Singular ADMGs.}
The construction of Singular~I is described in the main text. For Singular~II, we first sample the free diagonal and bidirected entries of $\Omega$, keeping only positive-definite draws that satisfy $|\omega_{12}|,|\omega_{23}|\geq 0.2$. We then sample $\lambda_{34}$ and $\lambda_{24}$ freely and set
\[
\lambda_{13}
=\frac{\omega_{22}\omega_{33}-\omega_{23}^{2}}
       {\omega_{12}\omega_{23}}.
\]
The lower bounds on $|\omega_{12}|$ and $|\omega_{23}|$ keep the denominator away from zero, so this assignment is well defined and enforces the rank-deficient constraint
\[
\det(M_3)=\omega_{22}\omega_{33}-\omega_{23}^{2}
-\lambda_{13}\omega_{12}\omega_{23}=0.
\]
For Singular~III, we sample the free directed coefficients and the free
entries of $\Omega$, and then impose two zero constraints by setting
\[
\begin{aligned}
\omega_{24}&=-\lambda_{23}\lambda_{34}\omega_{22},\\
\omega_{44}&=-\lambda_{12}\lambda_{23}\lambda_{34}\omega_{14}
 +(\lambda_{23}\lambda_{34})^2\omega_{22}.
\end{aligned}
\]
These choices are equivalent to enforcing
\[
\begin{aligned}
m_1&=\lambda_{23}\lambda_{34}\omega_{22}+\omega_{24}=0,\\
m_2&=\lambda_{12}\lambda_{23}\lambda_{34}\omega_{14}
 +\lambda_{23}\lambda_{34}\omega_{24}+\omega_{44}=0.
\end{aligned}
\]
A proposed Singular~III point is rejected unless the resulting $\omega_{44}$
is positive and $\Omega\succ0$.

\paragraph{ABIC baseline.}
For the baseline we use the RICF-based ABIC procedure of
\cite{bhattacharya2021differentiable}, thresholding the final edge weights at $0.05$: any directed or bidirected edge whose fitted weight is at most $0.05$ in magnitude is dropped from the returned ADMG. To stay within our computational budget, we run a single initialization per data set rather than the five random restarts of the original work, and therefore perform no BIC-based selection across restarts. The bow-free constraint is imposed only where indicated in Table~\ref{tab:abic-bowfree}: Singular~II and Singular~III use it because their ground-truth graphs are bow-free, while the remaining experiments use the unconstrained ADMG class.

\begin{table}[t]
\centering
\caption{Use of the bow-free constraint for ABIC.}
\label{tab:abic-bowfree}
\begin{tabular}{lc}
\toprule
Experiment & Bow-free constraint \\
\midrule
Sachs & No \\
General randomized ADMGs & No \\
Singular I & No \\
Singular II & Yes \\
Singular III & Yes \\
\bottomrule
\end{tabular}
\end{table}

\paragraph{ADVI optimization.}
For each data set, we run ADVI from five structure-logit initializations that span sparse to dense initial graph tendencies; the five runs are trained independently. After optimization, we retain the candidates that can be converted to PAGs and select the one with the smallest
$\widehat{\mathrm{WBIC}}(\mathcal{G})$, evaluated through the ADVI estimator of
Eq.~\eqref{eq:wbic-advi} with $S=100$ Monte Carlo samples,
\begin{align*}
&\widehat{\mathrm{WBIC}}(\mathcal{G})
=\frac{1}{S}\sum_{s=1}^{S} nL_n(\theta^{(s)}), \quad \text{where}\\
&\theta^{(s)}=T_{\mathcal{G}}^{-1}\!\left(\mu^\star+\sigma^\star\odot\varepsilon_s\right),
\quad \varepsilon_s\sim\mathcal{N}(0,I),
\end{align*}
where $nL_n(\theta)=-\sum_{i=1}^{n}\log p(x_i\mid\theta,\mathcal{G})$ is the data-fit term of Eq.~\eqref{eq:wbic} and the samples $\theta^{(s)}$ are drawn from the fitted variational posterior $q_{\psi^\star}$. Since $\widehat{\mathrm{WBIC}}$ is a free energy, a smaller value corresponds to a higher approximate marginal likelihood, and we select the candidate minimizing it. The Gumbel temperature is held fixed at $\tau_{\mathrm{gumbel}}=1$ throughout training. Writing $T$ for the number of optimization steps, the acyclicity-penalty weight follows a clipped exponential schedule
\[
\begin{aligned}
r&=\lambda_{\mathrm{final}}/\lambda_{\mathrm{init}},\\
\lambda_{\mathrm{acyc}}(t)
 &=\min\!\left\{\lambda_{\mathrm{final}},
 \lambda_{\mathrm{init}}r^{2t/T}\right\},
\end{aligned}
\]
so it reaches its final value after roughly half of the steps and stays
there. Finally, we threshold the soft graph to obtain a discrete ADMG; if its directed part is cyclic and thus cannot be converted to a PAG, we remove directed edges one at a time, starting with the one with the smallest soft-gate value in the detected cycle, until the graph is acyclic.

We use the following optimizer settings. For the general experiments:
$2{,}000$ steps, learning rate $0.01$, and $8$ Monte Carlo samples per update, with final acyclicity weight $10$ and both sparsity penalties set to zero. For every singular graph: $2{,}000$ steps, learning rate $0.005$, $15$ Monte Carlo samples per update, final acyclicity weight $10$, and sparsity penalties $\lambda_D=0.1$ and $\lambda_B=0$. For Sachs: $4{,}000$ steps, learning rate s$0.01$, $15$ Monte Carlo samples per update, final acyclicity weight $1000$, and sparsity penalties $\lambda_D=2$ and $\lambda_B=30$. The inverse temperature is $\beta=c/\log n$. We use the colder $c=4$ for the singular experiments, where distinguishing the over-parameterized graph from its submodel is hardest and the mean-field family most underestimates the posterior correlations along the singular set; outside these benchmarks we found the standard $c=1$ sufficient and use it for the general and Sachs experiments.

\end{document}